\pdfoutput=1

\documentclass[11pt]{article}

\usepackage[]{acl}

\usepackage{times}
\usepackage{latexsym}
\usepackage{multirow} 
\usepackage[T1]{fontenc}
\usepackage{graphicx}
\graphicspath{ {./img/} }
\usepackage[utf8]{inputenc}

\usepackage{microtype}
\renewenvironment{quote}{%
  \list{}{%
    \leftmargin 0.2cm   
    \rightmargin\leftmargin
  }
  \item\relax
}
{\endlist}

\title{Multi-Scales Data Augmentation Approach In Natural Language Inference For Artifacts Mitigation And Pre-Trained Model Optimization}

\author{Zhenyuan Lu \\
Department of Computer Science\\
University of Texas at Austin \\
  \texttt{zhenyuanlu@utexas.edu} \\}

\begin{document}

\maketitle
\begin{abstract}
  Machine learning models can reach high performance on benchmark natural language processing (NLP) datasets but fail in more challenging settings. We study this issue when pre-trained model learns dataset artifacts in natural language inference (NLI), the topic of studying the logical relationship between a pair of text sequences. We provide a variety of techniques for analyzing and locating dataset artifacts inside the crowdsourced Stanford Natural Language Inference (SNLI) corpus. We study the stylistic pattern of dataset artifacts in the SNLI. To mitigate dataset artifacts, we employ an unique {\it multi-scale data augmentation} technique with two distinct frameworks: a behavioral testing checklist at the sentence level and lexical synonym criteria at the word level. Specifically, our combination method enhances our model's resistance to perturbation testing, enabling it to continuously outperform the pre-trained baseline.

\end{abstract}

\section{Introduction}
\subsection{Background}

Natural Language Inference (NLI) is a fundamental subset of Natural Language Processing (NLP) that investigates whether a natural language sequence premise $p$ can infer (entailment), not imply (contradiction), or remain undetermined (neutral) with respect to a natural language sequence hypothesis $h$. \cite{maccartney2009,maccartney-manning-2008-modeling}. The challenge of NLI differs from that of other topics in that it highlights informal reasoning, lexical semantic knowledge, and the variety of language expression other than formal reasoning.

\subsection{Large-scale NLI Datasets}
To well-understand the semantic representation in NLI, and address the lack of large-scale materials, Stanford NLP groups have introduced the Stanford Natural Language Inference corpus \cite{emnlp2015}; later, Williams et al. from New York University have introduced the Multi-Genre Natural Language Inference (MultiNLI) corpus with collection of 433k sentence pairs modeled on the SNLI corpus \cite{N18-1101}. 

SNLI includes 570 k (550 k training pairs, 10 k development pairs, and 10 k test pairs) human-labeled sentence pairs categorized as entailment, contradiction, or neutral for training NLP models in NLI subjects. For the premise of the SNLI corpus, the researchers used captions from the Flickr30k corpus \cite{emnlp2015,young-etal-2014-image} from a collection of 160k crowdsourced captions. They utilized Amazon Mechanical Turk to gather hypotheses, gave crowd workers with a premise p, and requested them to construct three new hypotheses based on p, using one of the rules below \cite{emnlp2015, gururangan-etal-2018-annotation} and one of the examples from SNLI ({\textbf{Table~\ref{tab:SNLI}}}):

\begin{quote} 
  \noindent 
  {\bf Entailment}\hspace{9.2mm}$h$ is definitely true given $p$ \\
  \noindent 
  {\bf Neutral}\hspace{14.9mm}$h$ might be true given $p$ \\
  \noindent 
  {\bf Contradiction}\hspace{5mm}$h$ is definitely not true given $p$ \\
\end{quote}

\begin{table*}
  \centering
  \begin{tabular}{p{0.12\linewidth}@{\hspace{1cm}}p{0.1\linewidth}p{0.55\linewidth}}
    \hline
    \multirow{2}{*}{\bf Entailment}
    & Premise & A soccer game with multiple males playing.\\
    & Hypothesis & Some men are playing a sport.\\
    \hline
    \multirow{2}{*}{\bf Neutral}
    & Premise & An older and younger man smiling. \\
    & Hypothesis & Two men are smiling and laughing at the cats playing on the floor.  \\
    \hline
    \multirow{2}{*}{\bf Contradiction}
    & Premise & A man inspects the uniform of a figure in some East Asian country. \\
    & Hypothesis & The man is sleeping. \\
    \hline
  \end{tabular}
  \caption{SNLI sentence pairs with three labels: Entailment, Neutral, and Contradiction.}
  \label{tab:SNLI}
\end{table*}

\begin{table*}
  \centering
  \begin{tabular}{p{0.1\linewidth}@{\hspace{2cm}}p{0.55\linewidth}}
    \hline
    \bf Premise & A woman wearing all white and eating, walks next to a man holding a briefcase.\\
    \hline
    \bf Entailment & {\bf A person} eating. \\
    \bf Neutral & {\bf Two coworkers} cross pathes on a street.  \\
    \bf Contradiction & A woman eats ice cream walking down the sidewalk, and there is another {\bf woman} in front of her with a purse. \\
    \hline
  \end{tabular}
  \caption{Three sentence pairings from a single SNLI premise that highlight the artifacts produced by the annotation technique. In order to generate entailed hypotheses, it is usual practice to exclude gender or simplify the information. Sometimes, neutral hypotheses exclude gender information and just count the number of persons in the premise. Frequently, contradiction theories alter gender information.}
  \label{tab:artifacts}
\end{table*}

\subsection{Understand Dataset Artifacts}
Large-scale NLI datasets gathered through crowdsourcing are useful for training and evaluating NLU algorithms. Often, machine learning NLP models can achieve impressive results on these benchmark datasets. However, a new study has shown that NLP models can achieve incredibly high performance even without training on the premise corpus but on hypothesis baseline alone \cite{poliak-etal-2018-hypothesis}, although this strategy is difficult for models to learn theoretically. This issue known as dataset artifacts caused by human bias and spurious correlations occurred in the crowdsourcing-generated hypothesis, producing a stylistic pattern in the corpus \cite{gururangan-etal-2018-annotation}. 

For instance, a closer examination of the SNLI corpus ({\textbf{Table~\ref{tab:artifacts}}}) found that crowdsourcing frequently employs a few unexpected annotation strategies. {\textbf{Table~\ref{tab:artifacts}}} displays a single set of three pairings from SNLI: it is quite often to (1) change or omit gender information when generating entailment hypothesis, (2) neutralize the gender information and simply count the number of persons in the premise for neutral pairs, and (3) it is also typical to adjust gender information or only add negation content to create contradiction hypothesis. Consequently, NLP models trained on such datasets with artifacts tend to overestimate performance.

\section{Related Work}

\subsection{Detection on Dataset Artifacts}
Several strategies are developed to discover and analyze the dataset to understand its properties and identify any possible issues or dataset artifacts. Statistics such as the mean, standard deviation, minimum, and maximum values for each variable are computed to determine the shape of the data distribution and identify abnormalities. For instance, dataset artifacts may need to be addressed if the mean and standard deviation of a given corpus is significantly different from what statistical theory would predict \cite{gardner2021}. 

One research develops a collection of contrast examples by perturbing the input in various ways and testing the model's classification of these examples. This highlights the region in which the model may be making incorrect decisions, allowing for the possible correction of dataset artifacts that contribute to these inaccuracies \cite{gardner2020}. 

Another way is the checklist test, which involves creating a series of tests that cover a wide variety of common behaviors that NLP models should be able to handle, and then evaluating the model on these trials to see whether it has any artifacts \cite{ribeiro-etal-2020-beyond}. 

To evaluate model performance and identify and fix the observed dataset artifacts, Poliak et al. \cite{poliak-etal-2018-hypothesis} propose a hypothesis-only baseline by excluding the premise corpus from the dataset. Due to the unique method of producing NLI baseline datasets through crowdsourcing, related study suggests the model can also achieve good performance on hypothesis-only baseline dataset \cite{gururangan-etal-2018-annotation}.

\begin{figure*}[tp]
  \centering
  \includegraphics[width=\linewidth]{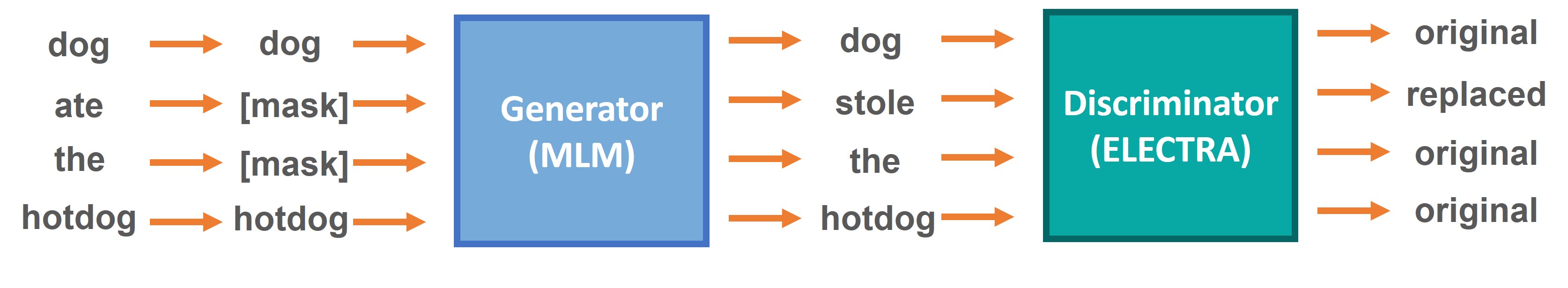}
  \caption{A summary of techniques for detecting replaced tokens. Token distributions can be created by any model, but often a small masked language model is trained in conjunction with the discriminator. Even though the models are built up similarly to a GAN, we train the generator using maximum likelihood instead of adversarially because GANs are often challenging to apply to text. After pre-training, we discard the generator and just alter the discriminator (the ELECTRA model) for subsequent jobs.}
  \label{fig:electra}
\end{figure*}

Other studies set up adversarial examples that challenge the NLI model's reliance on syntactic heuristics \cite{mccoy-etal-2019-right} or identify a set of universal triggers inserted into sentences to cause NLI model makes incorrect prediction \cite{wallace2019}. Similarly, some researchers create adversarial examples with challenges on lexical inferences \cite{glockner-etal-2018-breaking} and multiple choice questions requiring the NLI model to make challenging inferences \cite{zellers-etal-2018-swag}. Some studies employ adversarial examples to evaluate the NLI model's use of syntactic heuristics \cite{mccoy-etal-2019-right}, while others look for universal triggers that are inserted into sentences to cause the NLI model to make an incorrect prediction \cite{wallace2019}. Similarly, the NLI model is evaluated using adversarial situations that include lexical inference challenges \cite{glockner-etal-2018-breaking} and multiple-choice assessments \cite{zellers-etal-2018-swag}

\subsection{Address Dataset Artifacts}

There is more than one approach for dealing with the artifacts in the dataset. When the available training data for a natural language processing (NLP) model is insufficient, it can be supplemented with "adversarial datasets," which are similar to "adversarial challenge sets" in that they are generated by perturbing and transforming the original corpus on various levels to generate additional sets for NLP model learning \cite{morris2020textattack}. Another study presents ways for addressing two biases in the baseline dataset: contradiction - word bias and word - overlapping bias, by repeating training data and introducing synthetic data while applying a model - level debiases algorithm \cite{zhou-bansal-2020-towards}. Liu et al. describe a method for "inoculating" a model by fine-tuning it on a small, problem-specific dataset \cite{liu2019}, which helps the model learn to better handle the unique obstacles present in the dataset. One study collects and analyzes a huge number of human evaluations of samples in order to comprehend the difficulties and complexity of natural language inference. When comparing the capabilities of NLP models to that of 'collective' human intelligence, it is more appropriate to examine the models' ability to predict the whole range of human judgements, as opposed to only individual or majority opinions \cite{ynie2020chaosnli, xzhou2022distnli}.

\section{Our Implementation Details}

\subsection{Pre-Trained Model}

\begin{figure*}[tp]
	\centering
	\includegraphics[width=1\linewidth]{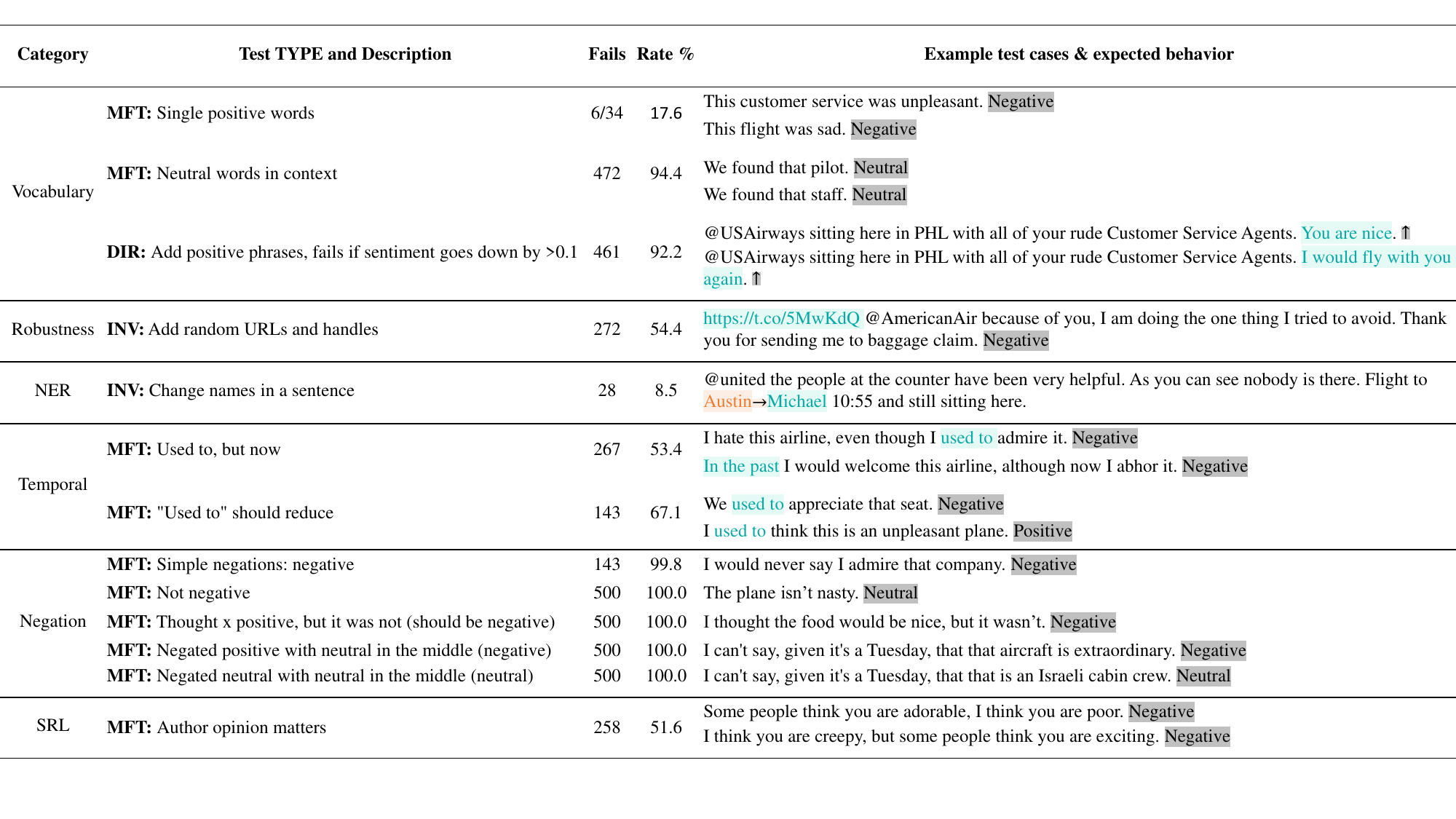}
    \caption{Checklisting fine-tuned ELECTRA-small model trained on SNLI dataset. Left channel shows the testing type and its description; middel channel presents the number of failed cases out of 500 testing cases, and the failure rate (\%); failure examples on the right channel.}
    \label{fig:checklist}
    \vspace{-10pt}
\end{figure*}

In this research, we implement the Efficiently Learning an Encoder that Classifies Token Replacements Accurately (ELECTRA, small) as our pre-trained model. ELECTRA is designed to improve the performance of BERT \cite{clark2020electra}. The method is similar to Noise-Contrastive Estimation (NCE) in that it trains a binary classifier to differentiate between real and fake data points \cite{pmlr-v9-gutmann10a}. On the other hand, ELECTRA may be seen as a contrastive learning model with a substantially scaled-up version of Continuous Bag-of-Words with negtative sampling \cite{Mikolov2013, clark2020electra}. Specifically, the purpose of contrastive learning is to embed augmented samples from the same sample near to each other while pushing away those that are not \cite{chen2020, smith-eisner-2005-contrastive,lu2022brief}. In the recent year, ELECTRA has gained popularity in the NLP area. It employs a training system distinct from BERT ({\textbf{Figure~\ref{fig:electra}}}). Multiple studies have demonstrated that it considerably increases performance on NLP tasks, such as NLI and question answering.

The ELECTRA-small model is a variant of the original ELECTRA model that is more compact and needs fewer computing resources during training and deployment. Despite its small size, the ELECTRA-small model is capable of achieving the same performance as the standard ELECTRA model for certain jobs. Therefore, it is preferable for use cases with limited computer resources.\\

\subsection{Dataset Artifacts Analysis}

{\it\textbf{Hypothesis-only SNLI}}

A hypothesis-only baseline, which only employs the hypothesis phrases of a dataset without the matching premise sentences, is one technique to investigate dataset artifacts in natural language inference. By analyzing the performance of a model trained on a hypothesis-only dataset, it is possible to identify potential artifacts or biases in the dataset that affects the model's performance.

Poliak et al. \cite{poliak-etal-2018-hypothesis} discovers that a hypothesis-only baseline model performed remarkably well on several natural language inference datasets, implying that the hypothesis sentences in these datasets included adequate information for the model to generate correct predictions. This finding implies that the hypothesis phrases in these datasets may contain redundant or unnecessary information that is not required for the job, potentially leading to dataset overfitting. Another research reveals that crowdsourcing commonly misclassifies specific types of annotations, such as negation and quantifiers, in natural language inference datasets \cite{gururangan-etal-2018-annotation}.

In our research, we propose to investigate dataset artifacts in natural language inference using a hypothesis-only SNLI dataset baseline. We will exclude all premise sentences from the dataset and only utilize hypothesis sentences for training and evaluation. To do this, the datasets.load dataset function from the datasets package will be used to load the SNLI dataset. Then, we will split the dataset into train and test sets and eliminate the premise sentences from the training and testing sets, leaving just the hypothesis sentences. Subsequently, the hypothesis-only training set will be employed to train the ELECTRA-small pre-trained model. We use the \verb|AutoModelForSequenceClassification| class to fine-tune the ELECTRA-small model.

Once the model is trained, we will evaluate its performance on the hypothesis-only testing set. We will also compare the performance of the hypothesis-only model to a model trained on the full SNLI dataset, with both premise and hypothesis sentences, to see how the removal of the premise sentences affects the model's performance. This approach aims to explore the potential artifacts and biases in the SNLI dataset by using a hypothesis-only baseline and to evaluate the performance of the ELECTRA-small model on this modified dataset.\\

\begin{table}
  \centering
  \begin{tabular}{lccp{0.4\linewidth}p{0.05\linewidth}@{\hspace{1.5cm}}p{0.1\linewidth}}
    \hline
    \textbf{Experiment} & \textbf{Accuracy} & \textbf{Dataset}\\
    \hline
    Hypothesis-Only & 69.76 & SNLI \\
    Hypothesis-Premise & 89.20 & SNLI \\
    \hline
  \end{tabular}
  \caption{SNLI accuracies on different experiments. Hypothesis-Only indicates the accuracy of the model only trained on hypothesis-Only SNLI dataset. Hypothesis-Premise is the accuracy which model trained on the SNLI datasets included premise and hypothesis.}
  \label{tab:hypotheseOnly}
\end{table}

\noindent{\it\textbf{Behavioral Test}}

To evaluate the efficacy of a model, we employ a suite of pre-defined tests as analytical metrics, and we've adopted the CheckList set from \cite{ribeiro-etal-2020-beyond} to do so. This technique employs the Minimum Functionality test (MFT), the Invariance test (INV), and the Directional Expectation test (DIR) as its three analytical metrics.
\begin{figure*}
	\centering
	\includegraphics[width=\linewidth]{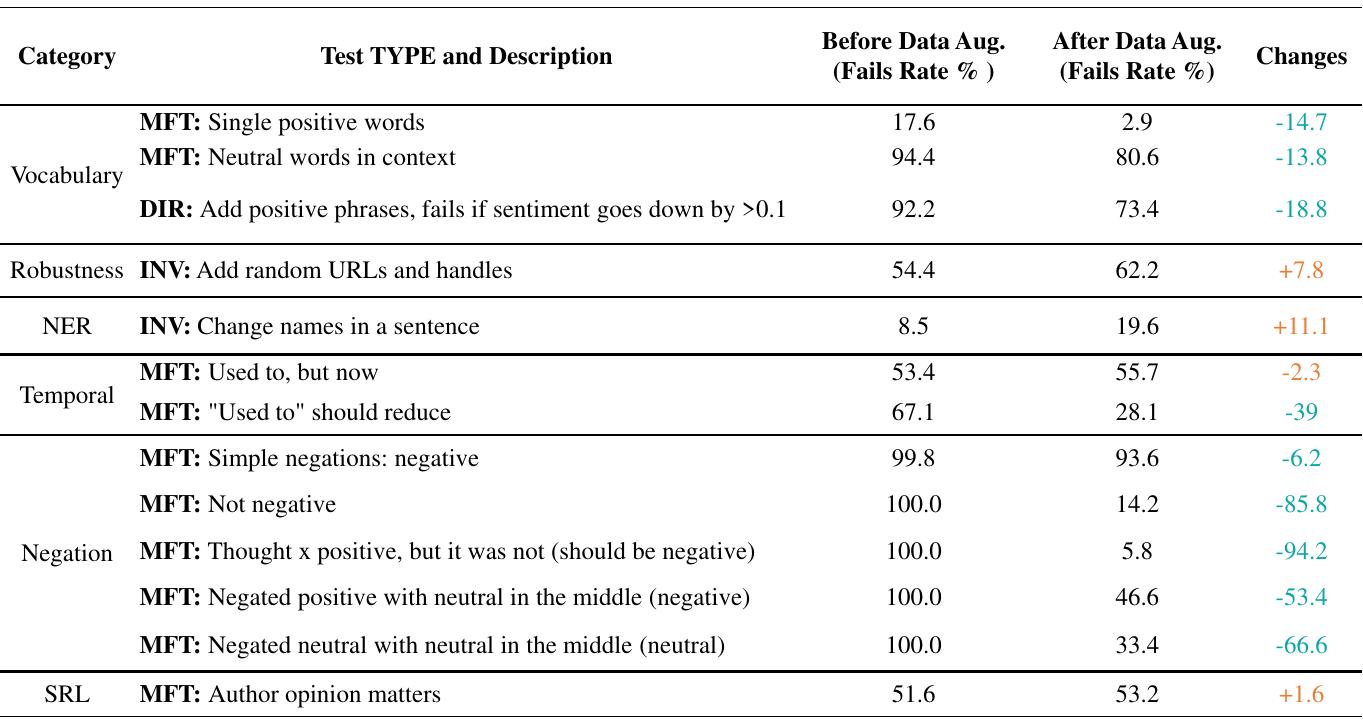}
    \caption{Checklisting fine-tuned ELECTRA-small model trained on SNLI dataset. Left channel shows the testing type and its description; right channel presents the improvment on the fails rate after data augmentation.}
    \vspace{-10pt}
    \label{fig:checklist_aug}
\end{figure*}
The MFT, e.g. negation test, is a tool for verifying a model's performance within a given set of constraints. This is achieved through the use of a library of generic instances annotated with generic labels. This helps find cases when models take short-cuts rather than mastering complicated inputs.

The INV test requires the model's prediction to be steady despite the introduction of input perturbations, e.g. name entity recognition (NER) testing.  For example, the NER capability for Sentiment modifies place names using specific perturbation functions. The results of this test are crucial for measuring the robustness of a model.

The DIR test is comparable to the INV test, however the label is anticipated to differ in a particular way. For instance, it is not expected that the sentiment of a sentence will become more positive if "I dread you." is added at the end of a comment "@JetBlue why won't YOU help them?! Ugh." \cite{ribeiro-etal-2020-beyond}, which generates monotonic shifts from negative to neutral mood. This test is useful for evaluating the consistency of a model.

The CheckList set approach provides a standardized mechanism for assessing the capabilities and behavior of NLP models. These analytic metrics give particular checks for many areas of a NLP model's performance, enabling a more thorough evaluation. \\

\begin{table}
  \centering
  \begin{tabular}{lccp{0.2\linewidth}p{0.05\linewidth}@{\hspace{1.5cm}}p{0.1\linewidth}}
    \hline
    \textbf{Dataset} & \textbf{Accuracy} & \textbf{Data Augmentation}\\
    \hline
    SNLI & 89.20 & No \\
    SNLI & {\textbf{89.79}} & CheckList and WordNet \\
    \hline
  \end{tabular}
  \caption{SNLI accuracies on different experiments. After data augmentation by using CheckList and WordNet strategies, our accuracy increases to 89.79 on SNLI dataset trained by pre-trained ELECTRA-small model.}
  \label{tab:snli_aug}
\end{table}

\noindent{\it\textbf{Adversarial Challenge}}

On our approach, we offer adversarial challenge as a second way for analyzing artifacts. In particular, we employ the Textattack framework to produce adversarial instances to attack our trained models using SNLI datasets in order to discover biases and spurious correlations \cite{morris2020textattack}. Textattack is a technique for evaluating the resilience of NLP models against adversarial attacks. This approach alters input data in such a fashion that a model makes erroneous predictions using a set of pre-defined adversarial instances. These adversarial examples are designed to be subtle and often go unnoticed by humans, yet they can cause a model to fail in certain ways.


\subsection{Dataset Artifacts Mitigation}

We propose a data augmentation strategy to mitigate the model's dataet artifacts based on the findings from the previous section. On our SNLI dataset, we employ two different scales of data augmentation approaches. The first is based on the CheckListing set \cite{ribeiro-etal-2020-beyond} and involves using a set of pre-defined rules to generate additional examples for a given dataset, such as adding positive phrases, altering names/locations, adding random URLs and handles, inserting temporal information, etc. This is connected to the preceding section's analysis of dataset artifacts; the extra step is to enrich data with such predefined collections. Using the checklist set, we generate two additional hypotheses for each initial premise in the SNLI dataset. This involves introducing label-preserving perturbations to the original corpus or expecting that the label will change dependent on the checklisting configuration or sentiment level. This will increase the quantity and variety of the dataset, which will help debias the original pattern established by crowdsourcing and decrease dataset artifacts. It can also aid in describing the model's core capabilities and behavior, leading to more precise predictions.

Another method we implemented is the word-level augmentation, WordNet list \cite{miller1995,fellbaum1998}. It utilize a lexical database of English words and their relationships. WordNet is designed to provide a rich thesaurus for word lexcial examples. It includes synonyms, antonyms, hyponyms, and hypernyms for each english word, as well as detailed definitions and examples of usage.

To prevent overlapping augmentation from the checkList set, we simply employ the synonym rule from the wordNet in our implementation. We scan each hypothesis phrase and produce two additional sentences by replacing certain synonym terms according to wordNet rules, while the premise corpus remains unchanged. This perturbation preserves the original dataset's label, as we only substitute synonym occurrences from each hypothesis sentence, which should not affect the original dataset's sentiment.

\section{Experiments}

\subsection{Stylistic Pattern from Crowdsourcing Generated Dataset}

Surprisingly, our model ablation research demonstrates that the hypothesis alone model achieves 69.76\% accuracy on the SNLI dataset ({\textbf{Table~\ref{tab:hypotheseOnly}}}). However, the results of a hypothesis-only model should not be so impressive. As a result of the stylistic pattern established by crowdsourcing, the dataset contains a high number of words that resemble the premise, indicating that the information in the hypothesis is redundant.

{\textbf{Figure~\ref{fig:checklist}}} shows that the pre-trained, finely-tuned ELECTRA-small model exhibits less-than-ideal performance across a variety of predefined tests, as determined by the checklisting tests. The examination of dataset artifacts indicates that the sentiment analysis model performs well in certain tests, including the vocabulary test with single positive words, where the failure rate was only 17.6\%, and the NER test of changing names in sentences, where the failure rate was only 8.5\%. However, when submitted to a robustness test that included random URLs and handles, the model failed 54.4\% of the time. This suggests that the model can detect sentences with a positive sentiment well, but poorly with a negative ones.

The model fail with a 99.8\% rate on tests involving basic negations and a 94.4\% rate on tests using neutral terms in context. The model also fail 100\% of tests that required more sophisticated negations, such as the test in which positive words were negated and neutral terms were sandwiched in the center.

The model failed with a failure rate of 53.4\% when evaluated with temporal phrases such as "used to" and "but now.". This raises concerns that the model has trouble understand context and adapting to shifting sentiments.

\subsection{Performance on SNLI Dataset}

Following the addition of more data, the model's accuracy increases to 89.79 percent, a 0.59 percent improvement over the fine-tuned baseline model ({\textbf{Table~\ref{tab:snli_aug}}}). However, the multi-scale data augmentation does help with artifacts by reducing the failure rate for vocab+POS and negation tests, both of which use predominantly positive and neutral pairings ({\textbf{Figure~\ref{fig:checklist_aug}}}). "MFT: Thought x positive, but it was not (should be negative)" and "MFT: Not negative" have seen the greatest falure rate reductions (-94.2 and -85.7 percentage points, respectively).

However, when the data augmentation does not cover all the dataset artifacts in the SNLI dataset, most of the negative challenges remain and the failure rate even increase, such as "INV: Change names in a sentence" which reach up to +11.1\% more compared to the model before data augmentation.

\section{Conclusions and Future Work}

To address this issue, we introduce a hybrid approach to identifying and mitigating SNLI dataset artifacts by combining a fine-tuned, pre-trained ELECTRA-small model. On the augmented SNLI dataset, its performance is superior to that of the baseline models. Yet there is a great deal of room for development. The first issue is dealing with the SNLI dataset, which has a heavily created pattern due to the crowdsourcing attempt, and finding the artifact with the biggest influence on the dataset.  The second fureture work can addresse how to optimize the architecture beyond the ELECTRA model. Contrastive learning among original and enhanced instances can be the foundation for enhancing performance based on a contrastive estimated loss, which can reduce the loss among instance bundles within the same premise group.



\bibliography{acl_latex}

\end{document}